%% file: arxiv_main.tex
\documentclass[11pt, copyright, logo, secondlogo]{arxiv_lark/LARK_Lab_Arxiv_Template/template/lark}
\usepackage[authoryear, sort&compress, round]{natbib}
\usepackage{bbm}
\usepackage{enumitem}
\setlist{nosep}
\usepackage{xspace}
\usepackage{graphicx}
\usepackage{tabularx}
\usepackage{booktabs}
\usepackage{subcaption}
\usepackage{xcolor}
\usepackage{hyperref}
\usepackage{verbatim}
\usepackage{float}
\usepackage{cleveref}
\usepackage{wrapfig}
\usepackage{makecell}
\usepackage{multirow}
\usepackage{fancyvrb}
\usepackage[normalem]{ulem}
\usepackage{url}
\usepackage{tcolorbox}
\usepackage{listings}
\tcbuselibrary{listings,skins,breakable}

\usepackage{amssymb}
\usepackage{algorithm}
\usepackage{algpseudocode}

\input{arxiv_lark/LARK_Lab_Arxiv_Template/math_commands.tex}

\graphicspath{{arxiv_lark/LARK_Lab_Arxiv_Template/}{figures/}}

\newcommand{\method}{\textsc{Switch}\xspace}
\newcommand{\swi}{\texttt{<swi>}\xspace}
\newcommand{\swiend}{\texttt{</swi>}\xspace}
\newcommand{\latent}{\texttt{<latent>}\xspace}

\title{Demystifying Hidden-State Recurrence: Switchable Latent Reasoning with On-Policy Reinforcement Learning}

\author[1]{Jiayu Yang$^{*}$}
\author[1]{Chao Chen$^{*}$}
\author[1]{Shengen Wu$^{*}$}
\author[2]{Yinhong Liu}
\author[3]{Yuxuan Fan}
\author[1]{Lujundong Li}
\author[1,4]{Songning Lai}
\author[1,5]{Chengwei Qin$^{\dag}$}
\author[1,5]{Zhijiang Guo$^{\dag}$}
\affil[1]{HKUST(GZ)}
\affil[2]{University of Cambridge}
\affil[3]{NTU}
\affil[4]{JoinQuant}
\affil[5]{HKUST}
\correspondingauthor{Zhijiang Guo (zhijiangguo@hkust-gz.edu.cn), Chengwei Qin (chengweiqin@hkust-gz.edu.cn)}
\githubpage{https://github.com/LARK-AI-Lab/SWITCH}
\modelweight{https://huggingface.co/LARK-Lab/SWITCH-Phase3-GRPO-LoRA-Qwen3-8B}

\usepackage{etoolbox}
\patchcmd{\maketitle}{\abscontent}{%
  {\noindent\small\bfseries Accepted to Findings of EMNLP 2026\par\vspace{8pt}}%
  \abscontent
}{}{\PackageError{acceptance-note}{Could not add acceptance note}{}}

\begin{abstract}

Latent chain-of-thought compresses reasoning through continuous hidden-state recurrence, but existing approaches are difficult to optimize with standard on-policy reinforcement learning (RL) and hard to analyze mechanistically. We introduce \textbf{\method}, a switchable latent reasoning framework that addresses both challenges with two discrete boundary tokens: \swi enters latent reasoning and \swiend exits it. These discrete anchors make policy ratios well-defined for standard on-policy RL while providing natural intervention points for mechanistic analysis. We train \method with a visible-to-latent curriculum and Switch-GRPO, which propagates gradients through recurrent latent computation. \method consistently outperforms prior hidden-state-recurrence approaches at similar model scales. Mechanistic analysis reveals that (i) \swi implements a localized, learned switching policy rather than a stylistic artifact; (ii) the latent step performs problem-specific, causally important computation; and (iii) this computation is concentrated in a single hidden-state transition upon entering latent mode. Together, these results show that hidden-state-recurrence latent reasoning can be both effectively optimized with on-policy RL and directly analyzed mechanistically, enabling us to study how RL reshapes internal reasoning computation. Code is available at \url{https://github.com/LARK-AI-Lab/SWITCH}.
\end{abstract}

\begin{document}
\maketitle

\section{Introduction}
\label{sec:intro}

Latent chain-of-thought (CoT) compresses the reasoning trace of a Large Language Model (LLM) by replacing visible text steps with
continuous latent steps. A natural realization of this idea is
\emph{hidden-state recurrence}, introduced by Coconut
\citep{hao2025coconut} and adopted by subsequent work
\citep{shen2025codi}: the latent step keeps computation inside
the LLM's own hidden space, feeding the previous step's last-layer
hidden state back as the next input embedding. The latent
computation thus runs in the LLM's existing representation space
and reuses the same forward pass that produces the surrounding
text, without introducing additional architectural components.
Two specific challenges have held the approach back. First,
on-policy reinforcement learning, now a standard tool for
aligning reasoning models with task rewards
\citep{deepseek2025r1,openai2024o1}---does not transfer cleanly
to the latent setting: latent positions emit no tokens and so
have no policy density, leaving methods such as GRPO undefined
inside the latent block. Existing systems therefore either skip
RL or run text-only training rollouts that diverge from the
inference-time decoder. Second, the latent computation is hard
to inspect: latent positions sit inside a continuous text
continuation with no token that an analyst can grip, leaving it
unclear whether the latent step performs task-relevant
computation or merely acts as an inert filler that the
surrounding text compensates for. We observe that both issues
share a common root cause: the absence of an explicit boundary
that marks where latent computation begins and ends.

This observation motivates our core idea: introduce a pair of
explicit boundary tokens that demarcate the latent block. The
model emits \swi to enter latent mode and \swiend to exit, with
hidden-state recurrence in between. The boundaries make latent
reasoning a learned, per-step decision---the model chooses
whether and when to invoke it---which is what \emph{switchable}
refers to in this paper. Two consequences follow: \swi and
\swiend are ordinary discrete tokens, so the GRPO ratio is
well-defined at every text position (latent positions simply
contribute no policy-gradient term); and the same boundaries
serve as anchors for analysis, letting us read $p(\swi)$, probe
the switch state from internal activations, and intervene on
specific latent hidden states.

The second affordance lets us address a recurring concern about
non-decoding ``thinking'' tokens
\citep{pfau2024letsthinkdotbydot,goyal2024pause}: that latent
positions might be non-functional placeholders the model has
learned to bypass, with the surrounding text doing the actual
work. Whether hidden-state-recurrence latents share this fate
has remained an open question, and the boundary tokens are what
let us answer it directly.

\begin{figure*}[t]
  \centering
  \includegraphics[width=0.92\textwidth]{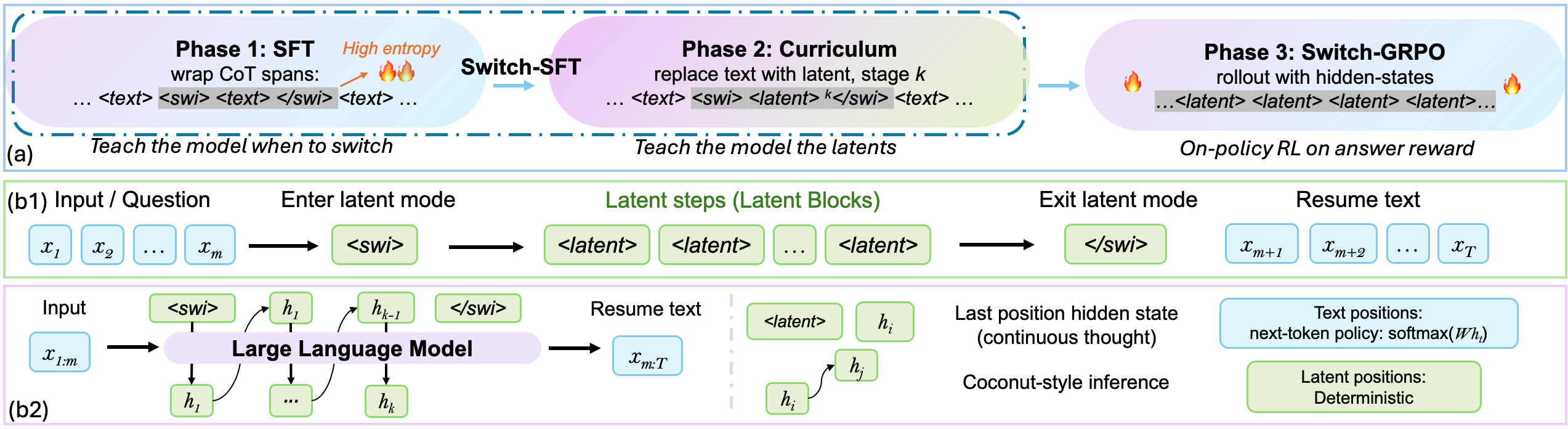}
  \caption{\textbf{\method overview.}
  \textbf{(a) Training.} Three phases turn a Qwen3 base into a
  switchable latent reasoner: SFT to wrap high-entropy CoT spans
  in \swi/\swiend, a curriculum that gradually replaces text
  inside those spans with \latent positions (jointly,
  \emph{Switch-SFT}), and Switch-GRPO for on-policy RL on the
  answer reward.
  \textbf{(b1) Inference token stream.} The model emits \swi to
  enter latent mode, runs a block of \latent steps, and emits
  \swiend to resume text decoding.
  \textbf{(b2) Hidden-state recurrence inside the block.} Each
  latent step's last-layer hidden state becomes the input
  embedding of the next \latent position (Coconut-style
  recurrence).}
  \label{fig:overview}
\end{figure*}

We package this idea as \method (Fig.~\ref{fig:overview}): the
model is trained in three phases---an SFT stage that wraps
high-entropy CoT spans in \swi/\swiend, a curriculum that
gradually replaces text inside \swi blocks with \latent positions
\citep[adapted from][]{hao2025coconut}, and a Switch-GRPO
optimizer that propagates gradients through the recurrent latent
computation \citep{sheng2025verl}. Our contributions are as
follows:

\begin{itemize}[leftmargin=*, itemsep=0pt, topsep=2pt, parsep=0pt]
\item \method addresses both challenges with one primitive:
      learned \swi/\swiend boundary tokens, paired with a
      Switch-GRPO optimizer, make on-policy RL well-defined and
      expose the latent computation to direct mechanistic
      analysis.
\item On MATH-500, \method reaches $\mathbf{79.3\%}$,
      $\mathbf{+25.7}$ points above the strongest Coconut-style
      baseline at the same scale; Switch-GRPO over the SFT-only
      checkpoint further halves the latent invocation rate while
      raising accuracy on invoked problems by
      $\mathbf{+12.6}$ points.
\item Mechanistic analysis through the boundary tokens yields
      three converging takeaways about the switch policy and the
      latent step's computation.
\end{itemize}

\section{Related Work}
\label{sec:related}

Latent CoT can be split by what a latent token is. Hidden-state
recurrence~\citep{hao2025coconut,shen2025codi} feeds the previous step's last-layer
hidden state back as the next input embedding; vocabulary mixtures~\citep{zhang2025softthinking, deng2026latentsft, deng2026latentgrpo, 
zheng2026softgrpo} instead sample a top-$k$ convex
combination of vocabulary embeddings via Gumbel-Softmax. Mixtures are samplable and admit direct policy gradients, which has motivated recent RL work to abandon hidden-state recurrence; we instead show the recurrence can be RL-trained (\S\ref{ssec:method-grpo}) and
verified mechanistically (\S\ref{sec:mechanistic}). Related
non-recurrent approaches include training-free switchable
inference on a frozen reasoning LLM~\citep{shi2026swireasoning}, adaptive test-time
compute that always emits visible thinking
\citep{snell2024scaling,chen2024overthinking}, and non-decoding
pause-style tokens
\citep{goyal2024pause,pfau2024letsthinkdotbydot,deng2024icot,tan2025colar}.
Detailed positioning of each line, together with the interpretability tools we apply in \S\ref{sec:mechanistic} including logit lens~\citep{nostalgebraist2020logitlens,belrose2023eliciting}, linear probing~\citep{tenney2019bertrediscovers,belinkov2022probingsurvey}, and causal activation interventions~\citep{meng2022rome,heimersheim2024patching}, is provided in Appendix~\ref{sec:appendix-related}. Other approaches extend latent reasoning in the multimodal setting. For instance, IVT-LR~\citep{chen2026reasoning} extends this idea by concatenating visual embeddings with hidden states to form a unified multimodal latent representation.
\section{Method}
\label{sec:method}

Training a hidden-state-recurrence latent reasoner is hard
because latent positions admit neither supervision targets nor a
sampling distribution, leaving both cross-entropy SFT and
standard policy-gradient RL undefined inside the block. \method
addresses this with a single primitive, the boundary tokens
\swi/\swiend, that gives every training stage a discrete handle
on the otherwise-continuous latent block. Three phases share it:
(i) SFT teaches the model when to emit \swi/\swiend; (ii) a
curriculum gradually replaces text inside the boundaries with
\latent positions while keeping the boundary signal intact; and
(iii) Switch-GRPO uses the same boundaries to make the GRPO
ratio well-defined at every text position, allowing on-policy
RL through trajectories that contain latent steps. We refer to
(i)~+~(ii) as \emph{Switch-SFT} and (iii) as \emph{Switch-GRPO};
full equations and algorithm boxes are in
Appendix~\ref{sec:appendix-grpo}
and~\ref{sec:appendix-algorithms}.

\subsection{Switchable Latent Reasoning}
\label{ssec:method-arch}

We extend the model's vocabulary with three special tokens: \swi
(enter latent), \swiend (exit latent), and \latent (latent
placeholder). At inference, the model decodes normally until it
emits \swi, runs at least $K_{\min}$ latent steps, and may then
emit \swiend to exit. The minimum dwell $K_{\min}$ is needed
because in Phase 2 every \latent run terminates with \swiend at a
fixed offset, and without forcing a few steps the trained model
exits in one; we explain why mechanistically in
\S\ref{ssec:mech-where}.

Following Coconut \citep{hao2025coconut}, the input embedding inside
a latent block is the previous step's last-layer hidden state:
\begin{equation}
\tilde{\bm{e}}_t \;=\;
\begin{cases}
E[x_t]      & x_t \neq \latent,\\[2pt]
\bm{h}_{t-1} & x_t = \latent.
\end{cases}
\label{eq:hsi}
\end{equation}
Because $\bm{h}_{t-1}$ depends on $\tilde{\bm{e}}_{1:t-1}$, this
is a recursion across latent positions: each latent step requires
its own forward pass through the model, with the previous step's
last-layer hidden state determining the next input embedding
(implementation details in Appendix~\ref{sec:appendix-impl}). At
text positions the next-token policy is the standard categorical
$\mathrm{softmax}(W\bm{h}_t)$. At latent positions
$\tilde{\bm{e}}_t$ is a Dirac mass and no token is sampled, which
is why hidden-state-recurrence latents admit no direct policy
density and what shapes the Switch-GRPO design below.

\subsection{Switch-SFT: Curriculum Study}
\label{ssec:method-sft}

Phase 1 and Phase 2 together form what we will call the
\emph{Switch-SFT} stage: a two-step supervised fine-tuning recipe
that takes a base model from visible CoT to a switchable latent
reasoner. Phase 1 teaches the model when to enter and exit the
latent block, and Phase 2 teaches it to do useful work inside
that block. We report the resulting checkpoint as ``after
Switch-SFT'' in \S\ref{sec:results} and use it as the
initialisation for Phase 3.

\paragraph{Phase 1: locating switch positions.}
Phase 1 trains the model to mark \emph{high-entropy} segments of
a visible CoT with \swi/\swiend. Following
SwiReasoning~\citep{shi2026swireasoning}, we measure
high-entropy positions on a mathematical CoT corpus
\citep{openr1math220k} as those where the base model's
next-token distribution has high Shannon entropy---intuitively,
positions where the model is uncertain about the next reasoning
step---and tag contiguous runs of such positions with the
boundary tokens. The annotated corpus is then used for standard
next-token cross-entropy supervised fine-tuning over the response
sequence (the prompt is masked from the loss as usual).

\begin{figure}[t]
  \centering
  \includegraphics[width=1\textwidth]{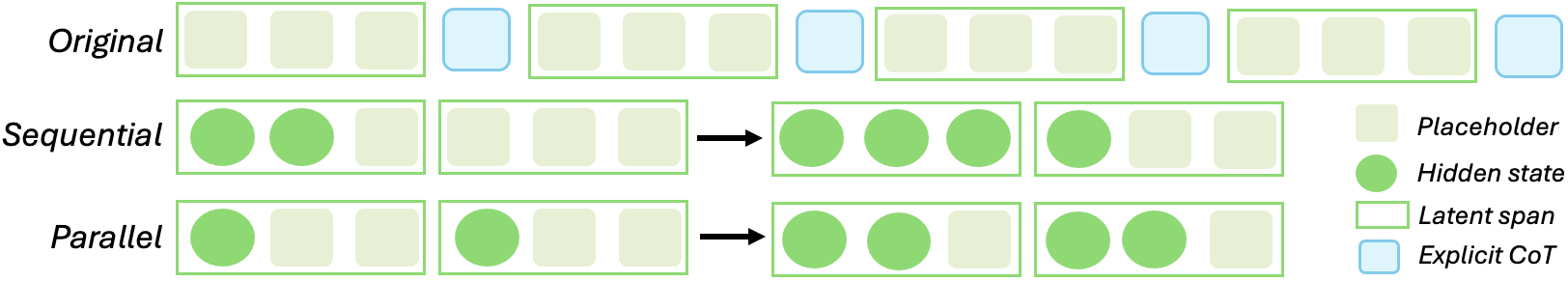}
  \caption{\textbf{Sequential vs.\ parallel curriculum
  schedules.} Hidden states (\emph{green circles}) replace
  text inside \swi-spans either one span at a time or in every
  span simultaneously across curriculum stages.}
  \label{fig:injection}
\end{figure}

\paragraph{Phase 2: latent curriculum.}
Phase 2 progressively replaces text inside \swi/\swiend blocks
with \latent positions while keeping \swi/\swiend in the loss,
so the model still has to learn when to enter and exit. A
one-shot replacement is too aggressive: with no prior experience
of latent computation, the model lets the block collapse into a
no-op. We compared two schedules
(Fig.~\ref{fig:injection}). Let $S_1,\ldots,S_M$ be the
\swi-spans of a sample and $|S_m|$ the text length of span $m$.
A \emph{sequential} schedule converts spans one at a time, so at
stage $k$ only the leftmost $k$ spans contain \latent positions.
A \emph{parallel} schedule, our default, converts every span
simultaneously and grows the per-span latent count:
\begin{equation}
n_m^{(k)} \;=\; c \cdot \min\!\bigl(k,\,|S_m|,\,K_{\max}\bigr),
\label{eq:curriculum}
\end{equation}
with $c\!=\!2$ and $K_{\max}\!=\!8$. \latent labels are masked,
so the loss applies to non-latent positions.

The parallel schedule is substantially better in our runs. Our
reading is that the sequential schedule keeps most of each
response inside the next-token-prediction distribution the base
model was pre-trained on, with only one span deviating at a
time, so the model can satisfy the loss without ever computing
in latent space. The parallel schedule pushes every span out of
that distribution at once and forces the model to produce
hidden states the surrounding text has to condition on. The
hyperparameter sweep and the head-to-head comparison are in
Appendix~\ref{sec:appendix-impl}.

\subsection{Switch-GRPO: Latent Exploring}
\label{ssec:method-grpo}

Phase 3 uses Group Relative Policy Optimization (GRPO)
\citep{shao2024deepseekmath} to improve correctness and
tag well-formedness. Two ingredients matter.

First, \emph{Switch-GRPO redefines what a rollout is}. Standard
GRPO assumes every rollout position
is sampled from a categorical token distribution and contributes
a policy density to the importance ratio. Hidden-state-recurrence
latent execution violates this assumption: \latent positions
emit no token, no sampling distribution, and no density.
Switch-GRPO resolves the conflict with two coupled changes.
\emph{(i) Rollout execution.} Rollouts run the same multi-pass
forward as the deployed decoder, so the trajectories the
optimiser sees at training time are exactly those produced at
inference. Standard text-only RL pipelines \citep{sheng2025verl}
silently bypass the latent step and train against a different
inference path. \emph{(ii) Likelihood factorisation.}
Hidden-state injection is deterministic given the preceding
text, so the rollout likelihood factors over text positions
only. The GRPO ratio is therefore well-defined at every \swi,
\swiend, and visible answer token; latent positions contribute
no policy-gradient term. Full equations are in
Appendix~\ref{sec:appendix-grpo}.

Second, \emph{the reward is correctness-dominant but
switch-aware}. We combine four terms in a weighted sum. A
$\pm 1$ correctness reward from \texttt{math-verify}
\citep{mathverify2025} dominates the signal. A $\pm 1$
tag-format reward enforces well-formed \swi/\swiend pairs. A
$\{0,1\}$ latent-usage reward pays out when a correct answer
uses \swi, encouraging the model to invoke the latent path
rather than fall back to plain text. The compression operating
point of \S\ref{ssec:results-pareto} adds an optional $[0,1]$
correctness-gated brevity term. The full reward formula, the
clipped surrogate loss, and the memory-segmented backward pass
are in Appendix~\ref{sec:appendix-grpo}.

\section{Experiments}
\label{sec:experiments}

\subsection{Experimental Setup}
\noindent\textbf{Model, Data and Benchmarks.}
All experiments, \method and every baseline in Table~\ref{tab:main}
alike, use Qwen3-8B \citep{qwen3} as the base model with three
special tokens (\swi, \swiend, \latent) added to the vocabulary.
Phases 1 and 2 use an annotated subset of OpenR1-Math
\citep{openr1math220k} whose high-entropy CoT sub-spans are wrapped
in \swi/\swiend following the SwiReasoning annotation pipeline
\citep{shi2026swireasoning}; Detailed training and hardware details are provided in Appendix~\ref{sec:appendix-impl}. For fair comparisons, we follow recent latent-CoT work
\citep{deng2026latentsft,deng2026latentgrpo} by using \textsc{MATH-500}
\citep{lightman2024letsverify,hendrycks2021math} and  \textsc{GSM8K}
\citep{cobbe2021gsm8k} as the benchmarks.

\noindent\textbf{Baselines.}
For Table~\ref{tab:main}, we re-implement every baseline on the
same base model under matched data and decoding settings,
following the protocols of CODI \citep{shen2025codi} and
Latent-GRPO \citep{deng2026latentgrpo}: a no-CoT direct-answer
baseline, a text-CoT SFT baseline trained on the same corpus, two
non-decoding ``thinking'' baselines (iCoT \citealp{deng2024icot},
Pause Tokens \citealp{goyal2024pause}), and three Coconut-style
latent reasoning methods that share the same hidden-state-injection
recurrence (Coconut \citealp{hao2025coconut}; CODI; CoLaR
\citealp{tan2025colar}). Re-implementation details are in
Appendix~\ref{sec:appendix-impl}.

\subsection{Main Results}
\label{sec:results}

This section is organized around the simplest version of the
question: does Switch-GRPO actually learn anything over the Coconut
curriculum alone, and if so, what does it learn? 
We first give the headline number against prior Coconut-style baselines, then compare the model immediately before
and after RL to isolate what changed.
A per-subject and per-difficulty breakdown is in
Appendix~\ref{sec:appendix-per-subject}. Finally, we show that the reward also gives users an explicit accuracy--length knob.

\paragraph{Headline Performance}
\label{ssec:results-headline}

Table~\ref{tab:main} reports \method and prior hidden-state-recurrence methods
(\S\ref{sec:related}), together with three standard non-latent
references (no-CoT direct answer, text-CoT supervised fine-tuning,
and the implicit/pause-token line). All baselines are evaluated on
the same base model under matched data and decoding settings,
so the comparison is apples-to-apples.

\method reaches $\mathbf{79.3\%}$ on MATH-500 and $\mathbf{89.2\%}$
on GSM8K, above all Coconut-style baselines under the
matched-base-model protocol. The ablation, accuracy--efficiency and mechanistic
analyses in \S\S\ref{ssec:results-sft-vs-rl}--\ref{sec:mechanistic}
are reported on a representative training run for which we have
full per-step training, decoding and intervention logs. 

\begin{table*}[t]
\centering
\small
\setlength{\tabcolsep}{5pt}
\begin{tabular}{llcccc}
\toprule
& & \multicolumn{2}{c}{\textbf{MATH-500}} & \multicolumn{2}{c}{\textbf{GSM8K}} \\
\cmidrule(lr){3-4}\cmidrule(lr){5-6}
\textbf{Method} & \textbf{Reasoning style} & Acc.\ & Tokens & Acc.\ & Tokens \\
\midrule
\multicolumn{6}{l}{\emph{Non-latent baselines}} \\
\quad No-CoT (direct answer)              & ---                     & 11.3 & 14   & 28.4 & 12   \\
\quad Text-CoT (SFT)                      & explicit                & 80.6 & 2\,079 & 88.6 & 1\,819 \\
\quad iCoT \citep{deng2024icot}           & internalised            & 24.8 & 9.6  & 60.4 & 9.5  \\
\quad Pause Tokens \citep{goyal2024pause} & non-decoding            & 14.6 & 14.7 & 33.7 & 13.5 \\
\midrule
\multicolumn{6}{l}{\emph{Coconut-style latent CoT}} \\
\quad Coconut \citep{hao2025coconut}      & hidden-state recurrence & 46.6 & 9.6  & 76.1 & 9.8  \\
\quad CODI \citep{shen2025codi}           & hidden-state recurrence & 48.3 & 10.2 & 76.4 & 9.9  \\
\quad CoLaR \citep{tan2025colar}          & hidden-state recurrence & 53.6 & 11.8 & 78.5 & 10.6 \\
\midrule
\quad \method (ours, after Switch-SFT)            & hidden-state recurrence & 66.7 & 1\,433 & 80.2 & 1\,249 \\
\quad \textbf{\method (ours, after Switch-GRPO)}  & hidden-state recurrence & \textbf{79.3} & \textbf{1\,721} & \textbf{89.2} & \textbf{1\,608} \\
\bottomrule
\end{tabular}
\caption{\textbf{Headline comparison on MATH-500 and GSM8K against
Coconut-style latent reasoning baselines.} All methods share a
common Qwen3-8B base model \citep{qwen3} under matched training
data and decoding settings; baseline numbers come from our own
re-implementations (Appendix~\ref{sec:appendix-impl}). ``Acc.''\
is accuracy (\%); ``Tokens'' is the average number of visible
(text) tokens per problem.}
\label{tab:main}
\end{table*}

\paragraph{What Does Switch-GRPO Add Over the Curriculum?}
\label{ssec:results-sft-vs-rl}

Table~\ref{tab:main} is the right number to report, but it does
not show \emph{where} the gain comes from. To separate the effect
of RL from the effect of the curriculum alone, we evaluate the
strongest curriculum-SFT checkpoint and the same checkpoint after
Switch-GRPO on the same MATH-500 set with the same $K_{\min}\!=\!4$
greedy decoding (Table~\ref{tab:sft-vs-rl}).

\begin{table}[t]
\centering
\small
\setlength{\tabcolsep}{4pt}
\begin{tabular}{lccc}
\toprule
\textbf{Stage} & \textbf{Latent acc.} & \textbf{Switch \%} & \textbf{Avg.\ tok.} \\
\midrule
After SFT (curriculum)         & 66.7 & 81 & 1\,433 \\
After Switch-GRPO              & \textbf{79.3} & 58 & 1\,777 \\
\bottomrule
\end{tabular}
\caption{Effect of Switch-GRPO on latent reasoning
ability. ``Latent acc.''\ restricts accuracy to test problems on
which the model emitted at least one \swi block.}
\label{tab:sft-vs-rl}
\end{table}

\begin{figure}[t]
  \centering
  \includegraphics[width=0.6\textwidth]{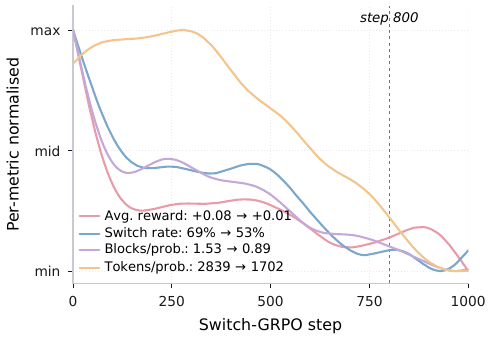}
  \caption{\textbf{Switch-GRPO training trajectory}, four metrics
  overlaid with per-metric min--max normalisation. Legend shows
  the actual start\,$\to$\,end values; the dashed line marks the
  reported step-$800$ checkpoint.}
  \label{fig:calibration}
\end{figure}

Two observations matter. Latent-conditional accuracy jumps by
$\mathbf{+12.6}$ points, attributable to RL alone since the
underlying weights, vocabulary, and decoding path are identical.
The switch rate drops from $81\%$ to $58\%$ at the same time. The
model has not learned to invoke latent reasoning indiscriminately;
it has learned to pick problems where the latent step pays off.
Figure~\ref{fig:calibration} shows the same calibration happening
continuously over the training run: as RL proceeds, latent
invocations per problem drop from $\sim\!1.5$ to $\sim\!1$ and
visible-token usage contracts from $\sim\!2900$ to $\sim\!1900$.
\S\ref{sec:mechanistic} returns to this calibration and
shows that the underlying switch policy is sharpened, not erased.

A per-subject and per-difficulty breakdown
(Appendix~\ref{sec:appendix-per-subject}) shows that the gain is
spread broadly: \method is strongest on algebraic and structurally
regular subjects (Algebra $88.7\%$, Prealgebra $80.5\%$, Number
Theory $79.0\%$) and accuracy degrades smoothly from $93.0\%$ at
level $1$ to $53.7\%$ at level $5$, with no cliff.

\begin{figure}[t]
  \centering
  \includegraphics[width=0.6\textwidth]{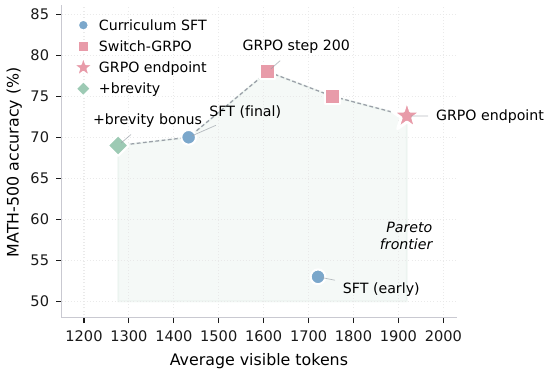}
  \caption{\textbf{Accuracy--efficiency operating curve} on the
  representative training run. MATH-500 accuracy vs.\ average
  visible tokens. The Switch-GRPO endpoint (star) sits at the
  high-accuracy end; the brevity-bonus operating point (diamond)
  trades a modest amount of accuracy for shorter outputs and zero
  max-length truncation. Dashed curve: empirical Pareto frontier.}
  \label{fig:pareto}
\end{figure}

\begin{figure}[t]
  \centering
  \includegraphics[width=0.6\textwidth]{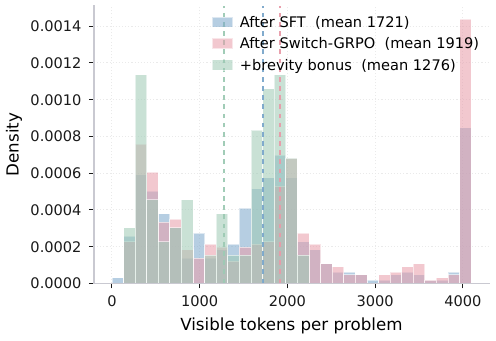}
  \caption{\textbf{Distribution of visible tokens per problem,}
  SFT vs.\ Switch-GRPO vs.\ +brevity. Dashed lines mark each
  checkpoint's mean.}
  \label{fig:token-dist}
\end{figure}

\noindent\textbf{An Accuracy--Efficiency Operating Curve}
\label{ssec:results-pareto} So far we have reported one operating point. By varying the
Switch-GRPO reward, in particular by adding a correctness-gated
brevity bonus (Appendix~\ref{sec:appendix-impl}), the user can
pick a point along an explicit accuracy--length curve
(Fig.~\ref{fig:pareto}). The brevity-bonus operating point trades
about three points of accuracy for $\sim\!33\%$ shorter outputs
and $0\%$ max-length truncation. The full visible token
distribution (Fig.~\ref{fig:token-dist}) shows that this is a
distributional shift rather than just a change of mean: the
brevity variant moves probability mass below the SFT median while
losing very few problems to the high-token tail. The matched
empirical CDF is in Appendix~\ref{sec:appendix-token-cdf}.

\section{How Does Latent Work in Reasoning?}
\label{sec:mechanistic}

The numbers in \S\ref{sec:results} say that Switch-GRPO produces
a better model than the curriculum-only SFT baseline. They do not
say \emph{why}. We use the explicit \swi/\swiend boundaries as
anchors to look at the trained model and answer three questions
in sequence: (Q1) does the model emit \swi with the localized
structure of a learned switching policy, (Q2) does the latent
step that follows contribute causally to the answer, and (Q3)
where inside the latent block does that contribution sit? We
answer Q1 with three observations
, Q2 with
a causal intervention, and Q3
with two complementary probes in this section.
Throughout we instrument two checkpoints from the same training
run: the curriculum-SFT checkpoint (\textbf{After SFT}) and the
post-RL endpoint (\textbf{After Switch-GRPO}).

\paragraph{\swi Is a Sharply Localised Boundary Token}
\label{ssec:mech-localization}

We teacher-force the model on the prefix immediately before an annotated \swi position and read out the next-token distribution,
using random non-boundary positions as a control. At annotated \swi positions, the model places \swi essentially at the top of the vocabulary (rank $\le 1.7$ on both checkpoints); at random non-boundary positions, it suppresses \swi by orders of magnitude
(rank $\sim\!10^3$, $p\sim\!10^{-3}$). The contrast is roughly
four orders of magnitude in $p(\swi)$ and three orders of
magnitude in rank. \swi is not a formatting artefact the model emits uniformly: it is a control action whose distribution cleanly separates reasoning boundaries from ordinary continuation positions (Table~\ref{tab:tf}).

\begin{table}[t]
\centering
\small
\setlength{\tabcolsep}{4.5pt}
\begin{tabular}{lcccc}
\toprule
\textbf{Event} & $H$ & $p(\swi)$ & rank & margin \\
\midrule
\multicolumn{5}{l}{\emph{After SFT}} \\
\quad \textsc{swi-start} & 0.203 & 0.847 & 1.13   & $+3.48$ \\
\quad random             & 0.068 & 0.003 & 1003.9 & $-21.9$ \\
\midrule
\multicolumn{5}{l}{\emph{After Switch-GRPO}} \\
\quad \textsc{swi-start} & 0.532 & 0.480 & 1.68   & $+0.08$ \\
\quad random             & 0.131 & 0.002 & 1127.9 & $-16.8$ \\
\bottomrule
\end{tabular}
\caption{\textbf{Teacher-forced switch statistics.} Annotated
\swi positions vs.\ random non-boundary positions, on both
checkpoints.}
\label{tab:tf}
\end{table}

\paragraph{The Switch-Window Has a Clean Spike.}
\label{ssec:mech-window}

\begin{figure}[t]
  \centering
  \includegraphics[width=0.55\textwidth]{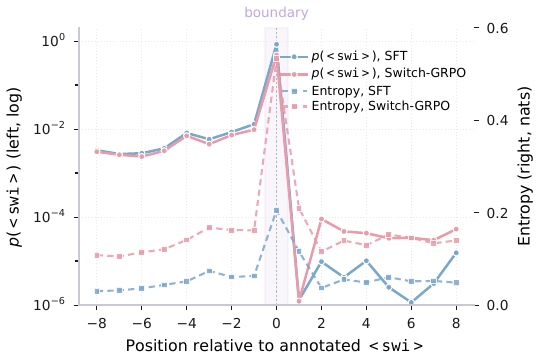}
  \caption{\textbf{Switch-window curves.} $p(\swi)$ (solid, left
  log axis) and entropy (dashed, right linear axis) at relative
  offsets $-8,\ldots,+8$ around annotated \swi positions. The
  spike at the boundary collapses by several orders of magnitude
  one token later. Switch-GRPO preserves the spike but softens
  its peak height; per-offset values at $-1, 0, +1$ in
  Table~\ref{tab:swi-window}.}
  \label{fig:swi-window}
\end{figure}

The next question is whether the model emits \swi only at the
boundary, or whether it emits \swi over a few-token region around
it. Figure~\ref{fig:swi-window} plots $p(\swi)$ at relative offsets
$-8,\ldots,+8$ around each annotated \swi position. The spike at
offset $0$ is followed by a collapse of several orders of magnitude
one token later, on both checkpoints; \swi is a boundary token,
not a stylistic tag spanning a window. The comparison between
checkpoints also sharpens the calibration story of
\S\ref{ssec:results-sft-vs-rl}: after SFT the peak sits at
$p(\swi)\!=\!0.85$ with large positive margin to the next token;
after Switch-GRPO it softens to $0.48$ with margin near zero, but
the contrast to the immediate neighbours stays at $\sim\!10^2$. RL
has not erased the switch policy. It has made the model less
aggressive at boundaries it is uncertain about, consistent with
the halved switch rate and the $+12.6$-point jump in
latent-conditional accuracy of Table~\ref{tab:sft-vs-rl}.

\begin{table}[t]
\centering
\small
\setlength{\tabcolsep}{6pt}
\begin{tabular}{lccc}
\toprule
\textbf{Checkpoint} & $p_{-1}$ & $p_{0}$ & $p_{+1}$ \\
\midrule
After SFT          & $1.3\!\times\!10^{-2}$ & $0.847$ & $2\!\times\!10^{-6}$ \\
After Switch-GRPO  & $9.7\!\times\!10^{-3}$ & $0.480$ & $2\!\times\!10^{-6}$ \\
\bottomrule
\end{tabular}
\caption{\textbf{Switch-window probabilities at the three central
offsets} $k\!\in\!\{-1, 0, +1\}$, with $p_k\!\equiv\!p(\swi)$.}
\label{tab:swi-window}
\end{table}

\paragraph{Switch State Is Linearly Decodable From Late Layers.}
\label{ssec:mech-probe}

\begin{figure}[h]
  \centering
  \includegraphics[width=0.55\textwidth]{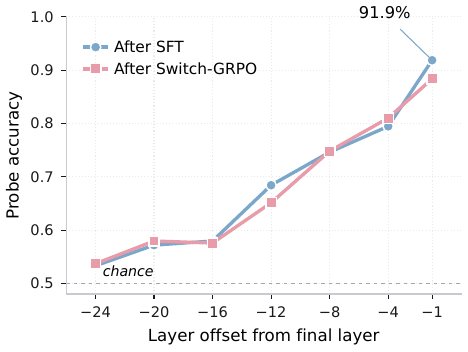}
  \caption{\textbf{Linear probe accuracy} for ``next token is
  \swi'' at seven layer offsets. Both checkpoints rise from near
  chance in the early layers to $\ge\!88\%$ at the last layer
  (exact numbers in Table~\ref{tab:probe}).}
  \label{fig:layer-probe}
\end{figure}

If the switch decision is being computed inside the model rather
than memorized at the output, we should be able to read it out of
the model's activations before the LM head. We fit a balanced
$\ell_2$-regularized logistic probe on the hidden state at seven
layer offsets, with the binary label ``next token is \swi''
(Fig.~\ref{fig:layer-probe}). The probe is near chance in the
early layers, becomes moderately predictive in the middle, and
reaches $91.9\%$ at the last layer (After SFT) or $88.4\%$ (After
Switch-GRPO): the classic ``feature emerges with depth'' picture
\citep{tenney2019bertrediscovers,belinkov2022probingsurvey}.
Switch-GRPO loses about three points at the last layer but keeps
the early- and mid-layer profile, consistent with the
softer-but-still-localized boundary policy.
\begin{table}[h!]
\centering
\small
\setlength{\tabcolsep}{6pt}
\begin{tabular}{lcc}
\toprule
\textbf{Layer offset} & \textbf{After SFT} & \textbf{After Switch-GRPO} \\
\midrule
$-24$  & 0.533 & 0.537 \\
$-20$  & 0.572 & 0.579 \\
$-16$  & 0.579 & 0.576 \\
$-12$  & 0.684 & 0.651 \\
$-8$   & 0.746 & 0.748 \\
$-4$   & 0.795 & 0.810 \\
$-1$   & \textbf{0.919} & \textbf{0.884} \\
\bottomrule
\end{tabular}
\caption{\textbf{Probe accuracy by layer offset} (balanced
\textsc{swi-start} vs.\ non-boundary classification).}
\label{tab:probe}
\end{table}

\noindent\textbf{Takeaway (Q1).}
\textit{\swi behaves as a learned switching policy, not a stylistic tag.}

\paragraph{The Latent Step Is Causally Doing Work.}
\label{ssec:mech-intervention}

\begin{figure*}[t]
  \centering
  \includegraphics[width=0.85\textwidth]{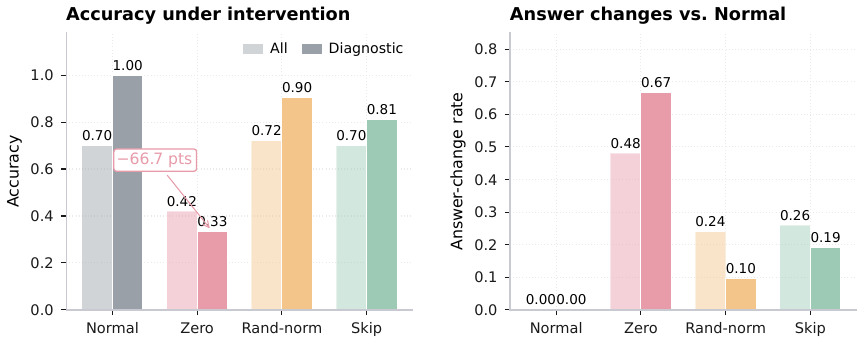}
  \caption{\textbf{Latent-state intervention.} Accuracy (left) and
  answer-change rate (right) under four intervention modes on the
  unrestricted MATH-500 set (light bars) and on the diagnostic
  subset (dark bars). \textsc{Zero} collapses diagnostic accuracy
  from $100\%$ to $33\%$ ($-66.7$ points); \textsc{Random-norm}
  and \textsc{Skip} are far less destructive. The latent
  computation is the specific hidden state of Eq.~\ref{eq:hsi},
  not just any non-zero perturbation.}
  \label{fig:intervention}
\end{figure*}

The three observations so far show that \swi is a switching
decision. They do not yet show that the latent step that follows
is doing useful work. To test that, we intervene on the injected
hidden state at every latent position
\citep{meng2022rome,heimersheim2024patching} and compare four
inference modes: \textsc{normal} (default generation),
\textsc{zero} (replace $\bm{h}_{t-1}$ in Eq.~\ref{eq:hsi} with the
zero vector), \textsc{random-norm} (replace it with a random
vector of the same norm), and \textsc{skip} (omit the latent step
and generate as if \swi had not been emitted). We run all four
modes on MATH-500 and report two summaries
(Fig.~\ref{fig:intervention}, Table~\ref{tab:intervention}): the
unrestricted average and a diagnostic subset of problems where
\textsc{normal} both used latent reasoning and answered correctly.
The diagnostic subset is the most informative for our question,
because on those problems the latent step is the only thing that
could matter.

\begin{table}[h]
\centering
\small
\setlength{\tabcolsep}{4pt}
\begin{tabular}{lcc|cc}
\toprule
& \multicolumn{2}{c|}{\textbf{All}} & \multicolumn{2}{c}{\textbf{Diagnostic}} \\
\textbf{Mode} & Acc.\ & $\Delta$ ans.\ & Acc.\ & $\Delta$ corr.\ \\
\midrule
\textsc{normal}      & 0.70 & 0.00 & 1.000 & $0.000$ \\
\textsc{zero}        & 0.42 & 0.48 & 0.333 & $\mathbf{-0.667}$ \\
\textsc{random-norm} & 0.72 & 0.24 & 0.905 & $-0.095$ \\
\textsc{skip}        & 0.70 & 0.26 & 0.810 & $-0.190$ \\
\bottomrule
\end{tabular}
\caption{\textbf{Latent-state intervention numbers.}
``$\Delta$ ans.''\ is the fraction of problems whose extracted
answer changes vs.\ \textsc{normal}; ``$\Delta$ corr.''\ is the
change in correctness on the diagnostic subset.}
\label{tab:intervention}
\end{table}

The contrast in the diagnostic subset is unambiguous. Zeroing
the latent state collapses accuracy from $100\%$ to $33.3\%$;
replacing it with a random vector of the same norm costs only
$9.5$ points; skipping the latent step costs $19.0$ points. Two
conclusions follow. The latent computation is not just
\emph{any} non-zero perturbation, since a same-norm random
vector is nearly harmless; and the latent step is not redundant
text in disguise, since skipping it costs twice as much as
random noise.

\noindent\textbf{Takeaway (Q2).}
\textit{The latent step performs causally important computation, not a
generic perturbation or redundant text in disguise.}

\paragraph{Where the Latent Computation Lives.}
\label{ssec:mech-where}

Q2 established that the latent step matters causally; we now
ask \emph{where} in the block the work happens. Two probes
converge on an answer. First, the logit lens
\citep{nostalgebraist2020logitlens,belrose2023eliciting}, applied
via $\mathrm{softmax}(W\bm{h}_t)$, returns \swiend as the top-$1$
token at every step, but at the \emph{first} latent step the
top-$k$ becomes more diffuse and problem-conditional (e.g.,
\texttt{inverse}, \texttt{arc}, \texttt{angle} on a trigonometry
problem). Second, $p(\swiend)\!\approx\!1$ at every latent step
regardless of correctness (Table~\ref{tab:exit}); without the
$K_{\min}$ constraint, the latent block would collapse to a
single hidden forward pass.

\begin{table}[t]
\centering
\small
\setlength{\tabcolsep}{6pt}
\begin{tabular}{lcccc}
\toprule
\textbf{Group} & step 1 & step 2 & step 3 & step 4 \\
\midrule
correct & 0.9998 & 1.0000 & 0.9904 & 1.0000 \\
wrong   & 1.0000 & 1.0000 & 0.9951 & 0.9999 \\
\bottomrule
\end{tabular}
\caption{\textbf{Exit probability $p(\swiend)$ inside latent,} stratified by whether the rollout is eventually correct.
The model is ready to exit immediately after entering.}
\label{tab:exit}
\end{table}

Together, the latent block reduces to a single hidden-state
transition on entry, kept from collapsing by $K_{\min}$ while
the curriculum's fixed-offset \swiend makes the remainder
exit-ready. This also explains the Q2 ordering: zeroing destroys
the transition that carries the reasoning, a same-norm random
vector preserves local geometry, and a \latent-skip walks past it.

\noindent\textbf{Takeaway (Q3).}
\textit{The work in the latent block is concentrated at a single
hidden-state transition on entry, kept alive by the $K_{\min}$
constraint.}

\section{Conclusion}
\label{sec:conclusion}

We presented \method, a switchable latent reasoning framework integrating a learned switch token, a three-phase curriculum, and a Switch-GRPO optimizer into hidden-state-injection models. Extensive experiments show that \method outperforms competitive baselines, while providing an adaptable accuracy--efficiency trade-off. Furthermore, its explicit boundary design allows for rigorous verification: the switch decision is highly localized and linearly decodable ($91.9\%$ probe accuracy), and causal analysis confirms the functional necessity of the latent reasoning steps. Overall, our work proves that recurrent latent CoT can be successfully optimized via RL and directly interpreted.

\section*{Acknowledgements}
This work is supported by funding from ModelBest and Guangdong Provincial Key Lab of Integrated Communication, Sensing and Computation for Ubiquitous Internet of Things (No. 2023B1212010007).

\section*{Limitations}
Our experiments are restricted to $8$B-parameter Qwen3 models and
to mathematical reasoning benchmarks (MATH-500 and GSM8K). We have
not yet evaluated \method on multi-domain reasoning or at larger
model scales, where the right balance between learned switching
and latent depth may differ. Switch-GRPO's gradient flows through the
text segments of each rollout only; latent positions contribute
via a frozen KV cache (Appendix~\ref{sec:appendix-impl}), so the
latent representation itself is shaped primarily by the Phase 2
curriculum rather than directly by the RL objective. Our mechanistic analysis is oriented
toward what is encoded by the model (switch boundaries, latent
causal effect) rather than at characterising failure modes; the
logit-lens decoding in \S\ref{ssec:mech-where} is
qualitative and should not be read as a faithful reconstruction of
the latent reasoning trajectory. A combined system in which the
latent token itself is also samplable, bridging hidden-state
recurrence and vocabulary-mixture latents, is a natural next step,
and a head-to-head comparison at matched scale and data is left to
future work.

\clearpage

\bibliography{custom}
\bibliographystyle{abbrvnat}

\clearpage

\appendix
\newpage
\section{Implementation Details}
\label{sec:appendix-impl}

\paragraph{Special tokens.}
We register three special tokens with IDs $151\,669$ (\swi),
$151\,670$ (\swiend), and $151\,671$ (\latent), resizing the Qwen3-8B
input/output embeddings from $151\,936$ to $151\,672$. The new
embeddings are initialised by anchoring to a content-neutral seed
token, which we found necessary to avoid rank-collapse with
$\mathcal{N}(0,\sigma)$ initialisation at $8$B scale.

\paragraph{Phase-1 SFT.}
We train Phase~1 with LoRA \citep[rank $32$, $\alpha\!=\!64$;][]{hu2022lora}
on all $\{q,\allowbreak k,\allowbreak v,\allowbreak o,\allowbreak\text{gate},\allowbreak\text{up},\allowbreak\text{down}\}$ projections,
together with the resized embeddings and the LM head. The standard
supervised cross-entropy loss reaches $0.098$ on the
annotated OpenR1-Math corpus.

\paragraph{Phase-2 curriculum.}
We use $c\!=\!2$ and $K_{\max}\!=\!8$ (up to $16$ \latent tokens per
span), a per-sample latent cap of $48$ to avoid OOM on samples with
many spans, and a curriculum-stage smoothing probability
$p_\text{unif}\!=\!0.1$. We sweep $k\!\in\!\{0,\ldots,8\}$ for three
epochs per stage, warm-starting each stage from the previous
checkpoint. The parallel replacement strategy is our default and
underlies all Phase-3 initialisations; the sequential warm-up
restricts the replacement at stage $k$ to the leftmost $k$ spans and
is reported in \S\ref{sec:appendix-ablation}.

\paragraph{Hardware.}
We train and evaluate on a single node of $8\!\times$NVIDIA H20
GPUs ($95$\,GB each) under PyTorch DDP. Switch-GRPO processes one
question per GPU with $G\!=\!5$ rollouts per question, applying
gradient updates atomically. We deliberately avoid vLLM-style
text-only rollout \citep{sheng2025verl} because it bypasses
hidden-state injection and breaks training and evaluation
alignment for latent reasoning.

\paragraph{Switch-GRPO hyperparameters.}
Group size $G\!=\!5$, clip threshold $\varepsilon_\text{c}\!=\!0.2$,
KL coefficient $\beta\!=\!10^{-3}$, learning rate $10^{-6}$, three
inner epochs per rollout. We use $\pi_{\theta_\text{old}}$ as the KL
anchor in place of a separate reference model, saving roughly
$18$\,GB/GPU. Rollouts use temperature $0.5$ in the main
configuration and $0.7$ in the compression-oriented operating point
of \S\ref{ssec:results-pareto}, $K_{\min}\!=\!4$ in the main
configuration and $K_{\min}\!=\!2$ in the compression operating
point, and $\text{max\_new\_tokens}\!\in\!\{2048,4096,6000\}$
depending on the configuration.

\paragraph{Segmented backward.}
Storing one autograd graph for $G$ rollouts, each containing many
latent passes interleaved with text, does not fit in
high-bandwidth GPU memory (HBM) at $8$B scale. We split every rollout at \swi/\swiend boundaries and process
the segments left-to-right, with a streaming key-value cache passing
between them. Latent segments run inside \texttt{torch.no\_grad()}
and store nothing; text segments run with gradient enabled,
contribute their term to the clipped surrogate loss of
Eq.~\ref{eq:grpo-loss} (below), and call \texttt{backward()}
immediately. Peak activation memory becomes that of one text
segment, so the only price we pay is that the latent positions
contribute to the gradient pass only through the KV cache they hand
off to the next text segment.

\paragraph{Reported metrics.}
For every checkpoint we report overall accuracy on each benchmark,
the switch rate (the fraction of test problems on which the model
emits at least one \swi block), and the average visible-token count
per problem. Latent forward passes are not counted as visible
tokens.

\paragraph{Compression-oriented operating point.}
For the shorter-output operating point of \S\ref{ssec:results-pareto}
we extend Switch-GRPO from the default checkpoint with an
additional brevity component $w_b\,r_\text{brev}$ in the reward
(Eq.~\ref{eq:reward-components}), $w_b\!=\!0.1$, $T_\text{lo}\!=\!800$,
$T_\text{hi}\!=\!2000$. The brevity bonus is gated on \emph{correct}
responses that used at least one \swi block, so the model is never
rewarded for producing short wrong answers. This operating point
reaches $69.0\%$ MATH-500 accuracy at $1\,276$ average visible
tokens with $0\%$ max-length truncation, versus the default
$72.6\%$ at $1\,919$ tokens with $18.4\%$ truncation, a
controllable Pareto trade-off rather than a separate ``best''
system.

\paragraph{Baseline re-implementations.}
For Table~\ref{tab:main} we re-implement every comparison method on
the same Qwen3-8B base model and OpenR1-Math training corpus. The
\emph{no-CoT} baseline emits only the final answer; the
\emph{text-CoT SFT} baseline trains the model on full visible CoT
without any switch token; \emph{iCoT} \citep{deng2024icot}
progressively internalises CoT steps; \emph{Pause Tokens}
\citep{goyal2024pause} insert non-decoding \texttt{<pause>} tokens
before the answer. For Coconut \citep{hao2025coconut}, CODI
\citep{shen2025codi}, and CoLaR \citep{tan2025colar} we follow each
paper's reference recipe but on Qwen3-8B: Coconut uses the
multi-stage curriculum with $c\!=\!2$ latent positions per text
token; CODI uses the single-stage self-distillation objective with
matched teacher/student; CoLaR uses a separate ``latent head'' that
predicts compressed embeddings.

\paragraph{Robustness.}
We wrap rollout and the gradient pass in
\texttt{try/except OutOfMemoryError} and propagate a
\texttt{survived\_indices} mask through the group-relative advantage
(Eq.~\ref{eq:grpo-adv} below), so a single long rollout does not
corrupt the advantage of an entire batch. Each question's $G$
rollouts are processed atomically (no gradient accumulation across
questions).

\section{Switch-GRPO Loss, in Full}
\label{sec:appendix-grpo}

\paragraph{Reward components.}
The reward terms described in \S\ref{ssec:method-grpo} of the main
body are defined as follows. Let $\hat y$ be the answer extracted from $o$ via
standard \texttt{\textbackslash boxed\{\}} parsing, $\equiv$ denote
mathematical equivalence judged by \texttt{math-verify}
\citep{mathverify2025}, $\mathbf{1}_\text{wf}$ indicate well-formed
\swi/\swiend tags, $\mathbf{1}_\text{used}$ indicate that at least
one \swi block is present, and $|o|$ the number of visible tokens:
\begin{align}
r_\text{corr} &= 2\,\mathbf{1}[\hat y \equiv y^\star] - 1, \\
r_\text{fmt}  &= 2\,\mathbf{1}_\text{wf} - 1, \\
r_\text{use}  &= r_\text{corr}\!\cdot\!\mathbf{1}_\text{wf}\!\cdot\!\mathbf{1}_\text{used}, \\
r_\text{brev} &= \mathrm{clip}\!\Bigl(\tfrac{T_\text{hi}-|o|}{T_\text{hi}-T_\text{lo}},0,1\Bigr) \notag\\
              &\quad \cdot \mathbf{1}[\hat y\!\equiv\!y^\star]\,\mathbf{1}_\text{used},
\label{eq:reward-components}
\end{align}
with $T_\text{lo}\!=\!800$ and $T_\text{hi}\!=\!2000$. The brevity
bonus is gated on correct responses that used \swi, so the model is
never rewarded for producing short wrong answers.

\paragraph{Group-relative advantages.}
Following \citet{shao2024deepseekmath} we use a trajectory-level
advantage shared across the rollout's text-positions,
\begin{equation}
A^{(i)} \;=\; \frac{R^{(i)} - \mu_R}{\sigma_R + \varepsilon},
\label{eq:grpo-adv}
\end{equation}
where $\mu_R$ and $\sigma_R$ are the mean and standard deviation of
$\{R^{(j)}\}_{j=1}^G$ and $\varepsilon\!=\!10^{-8}$.

\paragraph{Policy ratio.}
The per-text-position policy ratio at the frozen rollout prefix is
\begin{equation}
\rho^{(i)}_t(\theta) \;=\;
\frac{\pi_\theta(x^{(i)}_t \mid \tilde{\bm{e}}^{(i)}_{<t})}
     {\pi_{\theta_\text{old}}(x^{(i)}_t \mid \tilde{\bm{e}}^{(i)}_{<t})},
\label{eq:ratio}
\end{equation}
for $t\!\in\!\mathcal{T}_\text{text}^{(i)}$. The conditioning
$\tilde{\bm{e}}^{(i)}_{<t}$ is frozen from the rollout: we replay
the identical input-embedding sequence at the gradient pass, so the
same hidden-state injections that produced the reward are the ones
that backpropagate through the surrounding text.

\paragraph{Clipped surrogate loss.}
The full Switch-GRPO objective is the standard PPO-style clipped
surrogate plus a KL penalty, summed over text-positions of all
rollouts:
\begin{equation}
\mathcal{L}_3 \;=\;
-\,\mathbb{E}_q \!\!\!\!\!
\sum_{i,\,t\in\mathcal{T}_\text{text}^{(i)}}\!\!\!\!\!
\bigl[\,L^{(i)}_t - \beta\,\widehat{\mathrm{KL}}^{(i)}_t\,\bigr]
\Big/ N_\text{tok},
\label{eq:grpo-loss}
\end{equation}
with
\[
\begin{aligned}
L^{(i)}_t &= \min\!\bigl(\rho^{(i)}_t A^{(i)},\\
&\qquad \mathrm{clip}(\rho^{(i)}_t,1{-}\varepsilon_\text{c},1{+}\varepsilon_\text{c})A^{(i)}\bigr),
\end{aligned}
\]
$N_\text{tok} = \sum_i |\mathcal{T}_\text{text}^{(i)}|$,
$\varepsilon_\text{c}\!=\!0.2$, $\beta\!=\!10^{-3}$, and the unbiased
KL estimator
$\widehat{\mathrm{KL}}^{(i)}_t = \rho^{(i)}_t - 1 - \log \rho^{(i)}_t$
\citep{schulman2017ppo}. We use $\pi_{\theta_\text{old}}$ both as
the importance-sampling reference and as the KL anchor, removing
the need for a separate frozen reference model.

\section{Per-Checkpoint Trajectory of \method}
\label{sec:appendix-sweep}

The trajectory we use throughout the main body for ablation
(\S\ref{ssec:results-sft-vs-rl}), accuracy--efficiency
(\S\ref{ssec:results-pareto}), and mechanistic
(\S\ref{sec:mechanistic}) analyses is taken from a representative
training run for which we have full per-step training, decoding, and
intervention logs. Table~\ref{tab:appendix-sweep} reports its
MATH-500 numbers across the curriculum-SFT stages and Switch-GRPO
optimizer steps. The headline numbers reported in Table~\ref{tab:main}
($79.3\%$ MATH-500, $89.2\%$ GSM8K) come from our strongest
end-to-end Switch-GRPO run produced by the same pipeline; the
trajectory in Table~\ref{tab:appendix-sweep} provides the matched
diagnostic context.

\begin{figure*}[h]
  \centering
  \includegraphics[width=0.96\textwidth]{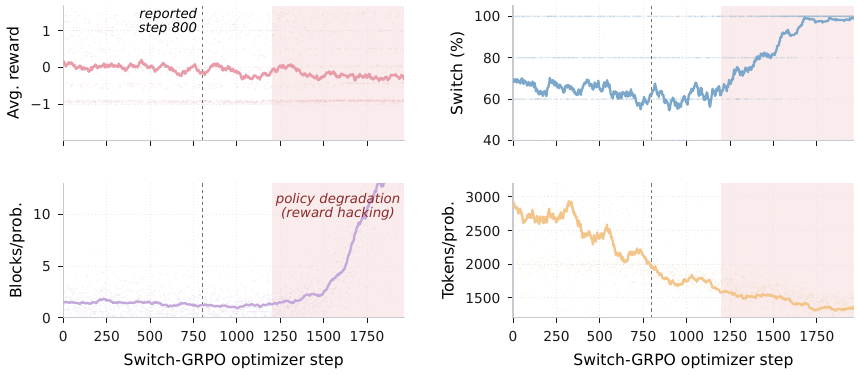}
  \caption{\textbf{Full Switch-GRPO trajectory} ($1{,}964$ optimizer
  steps) on the representative training run. Light scatter is raw
  per-batch metrics; coloured lines are $80$-step rolling means.
  The pink-shaded region marks the late-training reward-hacking
  regime ($\gtrsim\!\text{step}\ 1{,}200$); the dashed vertical line
  marks the reported step-$800$ checkpoint, our early-stopping point.
  The main-body Fig.~\ref{fig:calibration} shows the same panels
  cropped to the stable $0$--$1{,}000$ window.}
  \label{fig:calibration-full}
\end{figure*}

\paragraph{Full training trajectory.}
Figure~\ref{fig:calibration-full} extends the main-body
Fig.~\ref{fig:calibration} from step $1{,}000$ to the entire
$1{,}964$-step run, including the late-training region (shaded) in
which the policy enters a reward-hacking regime: switch rate climbs
to $100\%$, latent invocations per problem explode from $\sim\!1$ to
$\sim\!13$, and average reward drifts downward as the model invokes
\swi without converting the extra latent compute into correct
answers. We early-stop at step $800$, the operating point reported
throughout the main body; in the appendix view the dashed vertical
line and the shaded region together justify this choice.

\begin{table}[h]
\centering
\small
\setlength{\tabcolsep}{4pt}
\begin{tabular}{lccccc}
\toprule
\textbf{Checkpoint} & $K_{\min}$ & Acc.\ & Lat.\ acc.\ & Swi\,\% & Tok.\ \\
\midrule
\multicolumn{6}{l}{\emph{After SFT (curriculum only)}} \\
\quad stage 8       & 0 & 53.0 & 46.1 & 80 & 1721 \\
\quad stage 6       & 4 & 70.0 & 66.7 & 81 & 1433 \\
\midrule
\multicolumn{6}{l}{\emph{After Switch-GRPO}} \\
\quad step 200      & 4 & 78.0 & 74.6 & 67 & 1609 \\
\quad step 600      & 4 & 75.0 & 66.1 & 62 & 1753 \\
\quad \textbf{step 800} & 4 & \textbf{72.6} & \textbf{67.8} & 58 & 1919 \\
\midrule
\multicolumn{6}{l}{\emph{+ brevity bonus}} \\
\quad step 1000     & 4 & 69.0 & 75.4 & 57 & 1276 \\
\bottomrule
\end{tabular}
\caption{\textbf{Full per-checkpoint MATH-500 trajectory of the
representative \method training run.}
``After SFT'' refers to curriculum-only checkpoints (no RL);
``After Switch-GRPO'' refers to RL post-training from the strongest
SFT checkpoint with the default correctness + format + latent-usage
reward; ``+ brevity bonus'' adds the compression-oriented brevity
reward term (\S\ref{sec:appendix-impl}). The bolded row is the
post-RL endpoint of this representative run; the strongest
end-to-end Switch-GRPO run produced by the same pipeline reaches
$79.3\%$ MATH-500 / $89.2\%$ GSM8K (Table~\ref{tab:main}).}
\label{tab:appendix-sweep}
\end{table}

\section{Algorithm Boxes}
\label{sec:appendix-algorithms}

\begin{algorithm}[h]
\small
\caption{\texttt{CoconutSwiModel} forward.}
\label{alg:forward}
\begin{algorithmic}[1]
\Require tokens $x_{1:T}$, latent mask $L\!\in\!\{0,1\}^T$
\Ensure hidden states $\bm{h}_{1:T}$, logits $\bm{\ell}_{1:T}$
\State partition $1\!:\!T$ into maximal constant-$L$ segments $S_1,\ldots,S_M$
\State $\mathcal{K}\gets\emptyset$
\Comment{streaming KV cache}
\For{$m = 1,\ldots,M$}
  \State $\bm{e}_t \gets E[x_t]$ if $L_t\!=\!0$ else $\bm{h}_{t-1}$,\;\;$t\!\in\!S_m$
  \State $(\bm{h}_{S_m},\mathcal{K}) \gets f_\theta(\bm{e}_{S_m};\,\mathcal{K})$;\;\;
         $\bm{\ell}_{S_m}\gets W\bm{h}_{S_m}$
\EndFor
\State \Return $\bm{h}_{1:T},\,\bm{\ell}_{1:T}$
\end{algorithmic}
\end{algorithm}

\begin{algorithm}[h]
\small
\caption{One Switch-GRPO step.}
\label{alg:switch-grpo}
\begin{algorithmic}[1]
\Require prompt $q$, gold $y^\star$, group size $G$, clip
$\varepsilon_\text{c}$, KL coef $\beta$, min latent dwell $K_{\min}$
\State \textsc{// Rollout (no\_grad), real hidden-state injection}
\For{$i=1,\ldots,G$}
  \State $o^{(i)} \gets$ \Call{generate\_rl}{$q,\,\theta_\text{old},\,K_{\min}$}
  \Comment{Eq.~\ref{eq:hsi}}
  \State store $\mathcal{T}_\text{text}^{(i)},\;\tilde{\bm{e}}^{(i)}_{1:T_i}$
  \State $R^{(i)} \gets R(o^{(i)},q,y^\star)$ \Comment{Eq.~\ref{eq:reward-components}}
\EndFor
\State $\{A^{(i)}\}_{i=1}^G \gets \mathrm{normalise}\{R^{(i)}\}$
\Comment{Eq.~\ref{eq:grpo-adv}}
\State \textsc{// Segmented backward, gradient through text only}
\For{$i=1,\ldots,G$}
  \State partition $o^{(i)}$ into segments $S^{(i)}_{1:M_i}$ at \swi/\swiend
  \State $\mathcal{K}\gets\emptyset$;\quad $\ell_\text{cum}\gets 0$
  \For{$m=1,\ldots,M_i$}
    \If{$S^{(i)}_m$ is a latent segment}
      \State run $f_\theta$ on $S^{(i)}_m$ in \texttt{no\_grad}, update $\mathcal{K}$
    \Else\Comment{text segment, gradient on}
      \State compute $\rho^{(i)}_t$ on $S^{(i)}_m$ \Comment{Eq.~\ref{eq:ratio}}
      \State $\ell_\text{seg}\gets$ contribution of $S^{(i)}_m$ to $\mathcal{L}_3$
      \Comment{Eq.~\ref{eq:grpo-loss}}
      \State $\ell_\text{seg}.\textsc{backward()}$;\;$\ell_\text{cum}\!+\!=\!\ell_\text{seg}.\text{detach}$
      \State update $\mathcal{K}$ with this segment's $(K,V)$
    \EndIf
  \EndFor
\EndFor
\State $\theta\gets\textsc{optim.step}()$;\;\;$\theta_\text{old}\gets\theta$
\State \Return $\ell_\text{cum}/N_\text{tok},\;\{R^{(i)}\}$
\end{algorithmic}
\end{algorithm}

\section{Visible-Token CDF}
\label{sec:appendix-token-cdf}

\begin{figure}[h]
  \centering
  \includegraphics[width=0.55\textwidth]{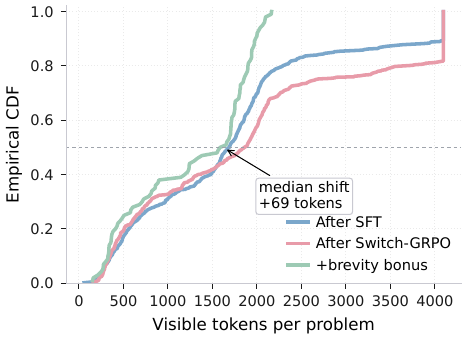}
  \caption{\textbf{Empirical CDF of visible tokens per problem.}}
  \label{fig:token-cdf}
\end{figure}

Figure~\ref{fig:token-cdf} reports the empirical CDF of visible
tokens per problem for the SFT baseline, the Switch-GRPO endpoint,
and the brevity-bonus operating point on MATH-500, complementing
the main-body histogram (Fig.~\ref{fig:token-dist}). The
brevity-bonus variant dominates the SFT baseline up to the median
while losing very few problems to the high-token tail.

\section{Mechanistic Analysis: Additional Details}
\label{sec:appendix-mech}

\paragraph{Probe details.}
The probe in Table~\ref{tab:probe} of the main body is a balanced
$\ell_2$-regularised logistic classifier ($C\!=\!1.0$) with an
$80\!:\!20$ train/test split. Reported numbers are test-set
accuracies.

\paragraph{Teacher-forced switch metrics.}
The numbers in Table~\ref{tab:tf} are computed by teacher-forcing
each instance on the prefix immediately before an annotated \swi
position, sampling random non-boundary positions per checkpoint as
a negative control, and reading out the entropy, $p(\swi)$, the
rank of \swi, and the log-margin to the top token. We use the same
MATH-500 evaluation problems for both checkpoints so the contrasts
are paired.

\paragraph{Switch-window window size.}
Figure~\ref{fig:swi-window} uses offsets $-8,\ldots,+8$ around each
annotated \swi position. The collapse one token after the boundary
($\sim\!2\!\times\!10^{-6}$ for $p(\swi)$, rank $\sim\!5\,000$) is
robust to window-size choice.

\paragraph{Probe details.}
We balance the binary classification dataset by sampling an equal
number of non-boundary positions per \textsc{swi-start} position,
and fit a logistic probe with default $\ell_2$ regularisation
($C\!=\!1.0$) and a single $80\!:\!20$ train/test split. Reported
numbers are test-set accuracies.

\paragraph{Intervention modes.}
For the diagnostic subset of Table~\ref{tab:intervention} we first
run \textsc{normal} inference and restrict to the problems where it
(i) produced at least one latent block and (ii) was graded correct
by \texttt{math-verify}. Each intervention mode is then run under
greedy decoding with $K_{\min}\!=\!4$ and the same
$\text{max\_new\_tokens}$ as the headline evaluation.

\section{Per-Subject and Per-Difficulty Visualisation}
\label{sec:appendix-per-subject}

\begin{figure*}[h]
  \centering
  \includegraphics[width=0.85\textwidth]{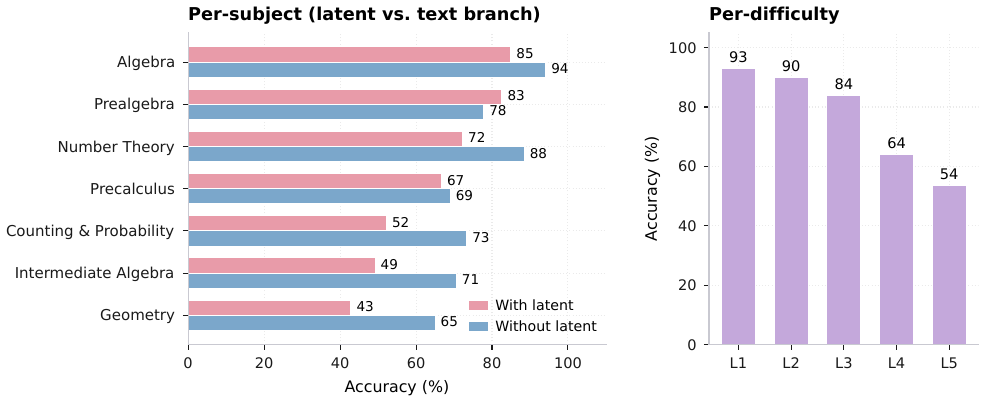}
  \caption{\textbf{Per-subject (left) and per-difficulty (right)
  MATH-500 accuracy} of \method after Switch-GRPO.}
  \label{fig:per-subject}
\end{figure*}

Table~\ref{tab:per-subject} reports \method's MATH-500 accuracy
split by subject and by difficulty level for the headline post-RL
checkpoint. Figure~\ref{fig:per-subject} additionally splits each
subject into ``with latent'' and ``without latent'' branches.

\begin{table}[h]
\centering
\small
\setlength{\tabcolsep}{8pt}
\begin{tabular}{lc}
\toprule
\textbf{Subject} & \textbf{Acc.\ (\%)} \\
\midrule
Algebra                 & 88.7 \\
Prealgebra              & 80.5 \\
Number Theory           & 79.0 \\
Precalculus             & 67.9 \\
Counting \& Probability & 60.5 \\
Intermediate Algebra    & 56.7 \\
Geometry                & 53.7 \\
\midrule
\textbf{Level 1}        & 93.0 \\
\textbf{Level 2}        & 90.0 \\
\textbf{Level 3}        & 83.8 \\
\textbf{Level 4}        & 64.1 \\
\textbf{Level 5}        & 53.7 \\
\bottomrule
\end{tabular}
\caption{\textbf{Per-subject (top) and per-difficulty (bottom)
MATH-500 accuracy} of \method after Switch-GRPO on the
representative training run.}
\label{tab:per-subject}
\end{table}

\section{Generation Trace Analysis}
\label{sec:appendix-gen}

\paragraph{Failure profile of wrong trajectories.}
We additionally stratify \method's post-RL trajectories by
correctness and by latent usage on MATH-500
(Table~\ref{tab:gen-analysis}). Wrong trajectories are substantially
longer than correct ones, both with and without latent usage, and
their switch decisions are slightly less confident
(entropy $0.717$ vs.\ $0.608$, $p(\swi)\!=\!0.669$ vs.\ $0.763$).

\begin{table}[h]
\centering
\small
\setlength{\tabcolsep}{4pt}
\begin{tabular}{lccc}
\toprule
\textbf{Group} & avg.\ tok.\ & avg.\ \swi & avg.\ latent steps \\
\midrule
correct, no \swi   &  959 & 0.0  & 0.0 \\
correct, with \swi & 1197 & 1.86 & 7.43 \\
wrong, no \swi     & 1729 & 0.0  & 0.0 \\
wrong, with \swi   & 1805 & 1.40 & 5.60 \\
\bottomrule
\end{tabular}
\caption{\textbf{Generation trace analysis} of \method (after
Switch-GRPO) on MATH-500, $K_{\min}\!=\!4$. Wrong trajectories are
substantially longer than correct ones in both branches, consistent
with the calibration story of \S\ref{ssec:mech-window}.}
\label{tab:gen-analysis}
\end{table}

\section{Ablations}
\label{sec:appendix-ablation}

\paragraph{Minimum latent dwell $K_{\min}$.}
We swept $K_{\min}\!\in\!\{0,2,4,8,16\}$ during inference; the
$K_{\min}\!=\!0$ setting collapses latent blocks to a single hidden
forward pass and reduces accuracy to $53.0\%$ (the
curriculum-SFT MATH-500 number reported in the trajectory of
Table~\ref{tab:appendix-sweep}), while $K_{\min}\!=\!4$ recovers the
training distribution and is our default
(\S\ref{ssec:mech-where} gives the mechanistic justification).

\section{Full Related Work}
\label{sec:appendix-related}

We give the more detailed treatment of related work promised in
\S\ref{sec:related}.

\subsection{Latent Chain-of-Thought Reasoning}

\paragraph{Hidden-state recurrence (Coconut, CODI).}
Coconut \citep{hao2025coconut} formulates continuous CoT by feeding
the previous step's last-layer hidden state back as the next input
embedding, so that an entire reasoning step happens in latent space
between two text tokens. The model is trained with a multi-stage
curriculum that progressively replaces explicit CoT tokens with
$k\!\cdot\!c$ latent positions. CODI \citep{shen2025codi} keeps the
same hidden-state-injection mechanism but replaces the curriculum
with a single-stage self-distillation objective: a teacher path with
full explicit CoT and a student path with a few continuous thoughts
share weights, and the student's hidden state at the answer-adjacent
token is aligned to the teacher's via an $L_1$ feature loss. Both
methods inherit the same latent geometry: each latent token has
input embedding equal to a previous-step hidden state, a
deterministic function of the input prefix. \method continues this
line.

\paragraph{Vocabulary-embedding mixtures.}
A more recent line redefines the latent token as a \emph{convex
combination} of vocabulary input embeddings. Soft-Thinking
\citep{zhang2025softthinking} uses the next-token softmax
probabilities as mixture weights, $s_t = \sum_{v \in
\mathcal{V}} p_t(v)\,E[v]$, so $s_t$ lies on the convex hull of the
vocabulary embeddings. Latent-SFT \citep{deng2026latentsft}
restricts this to a top-$k$ mixture and trains with stochastic
Gumbel-Softmax targets, reporting $2.7\!\times$--$5.5\!\times$
shorter chains than explicit SFT on six math benchmarks.
Latent-GRPO \citep{deng2026latentgrpo} explicitly contrasts itself
with Coconut, calling the latter ``early methods which directly
adopt the hidden state as the latent token,'' and proposes
vocabulary-superposition with one-sided Gumbel margins as a more
RL-friendly alternative. SofT-GRPO \citep{zheng2026softgrpo}
similarly adds Gumbel noise to logits to make Soft-Thinking
RL-trainable. The shared property of vocabulary mixtures is that
latent tokens are \emph{samplable} via Gumbel-Softmax and have a
tractable density, which is precisely what makes GRPO directly
applicable. We do not include the vocabulary-mixture line in our
headline comparison because it uses a different latent
representation; its results are not directly comparable, and
\method's contribution is orthogonal: we show that the original
hidden-state-recurrence representation can itself be made
RL-trainable.

\paragraph{Scale.}
Prior hidden-state-recurrence work has operated mainly at GPT-2 /
$1$--$2$B scale: Coconut's main experiments use GPT-2
\citep{hao2025coconut}, and CODI uses GPT-2 and LLaMA-1B
\citep{shen2025codi}. The Coconut paper reports a brief LLaMA-3-8B
probe in its appendix (improving GSM8K by $1.4$ points over the SFT
baseline), but without a tuned curriculum, a learned switching
token, or RL, which is the regime \method targets. The closely
related CoLaR \citep{tan2025colar} is also
hidden-state-recurrence-based but introduces a separate ``latent
head'' that predicts compressed embeddings.

\paragraph{Other compression approaches.}
Despite the strong generalization and reasoning abilities demonstrated by modern LLMs, the auto-regressive nature of LLMs imposes a significant computation burden, limiting their application in agent-based or edge scenarios. To address this, various compression approaches have emerged to improve the decoding efficiency~\citep{li2026too, yang2026accordion}. A simpler line inserts non-decoding ``thinking'' tokens without continuous-state feedback: pause tokens \citep{goyal2024pause},
filler tokens \citep{pfau2024letsthinkdotbydot}, and implicit-CoT internalisation \citep{deng2024icot}. These do not maintain explicit text reasoning at all, whereas \method preserves text CoT outside
\swi blocks so the visible trajectory remains interpretable.

\subsection{Switchable / Hybrid Reasoning}

The closest single work to ours is SwiReasoning
\citep{shi2026swireasoning}. It takes a frozen reasoning LLM and
dynamically switches between explicit decoding and a latent step
based on the entropy trend of the next-token distribution. A hard
\emph{switch budget} caps the number of mode changes per problem
\citep{chen2024overthinking,snell2024scaling}, yielding
$1.8$--$3.1$ accuracy points and $57$--$79\%$ token-efficiency
gains on math, STEM, coding, and general benchmarks.

\paragraph{Why we still train.}
SwiReasoning is training-free, so the latent step is performed by a
model that was not trained for it, and the location of switches is
fixed by an external entropy rule. \method instead \emph{learns} a
discrete switching token (\swi) and the latent dynamics inside it,
so both the entry point and the dwell of latent reasoning are
optimised end-to-end, in particular by RL in Phase~3.

\paragraph{Adaptive test-time compute.}
A separate line trains LLMs to spend test-time compute adaptively
\citep{snell2024scaling,chen2024overthinking}, but always emits the
extra ``thinking'' as text. \method brings the adaptive-compute
decision and the latent representation into a single trained model.

\subsection{Reinforcement Learning for Reasoning and Latents}

Group Relative Policy Optimization
\citep[GRPO;][]{shao2024deepseekmath} is the de-facto policy
optimizer for post-training reasoning LLMs and underpins
DeepSeek-R1 \citep{deepseek2025r1}; \citet{schulman2017ppo} provides
the PPO foundation.

\paragraph{RL for latent representations.}
Standard GRPO assumes the policy outputs a categorical distribution
over discrete tokens and that rollouts sample from it.
Vocabulary-mixture latents satisfy this through Gumbel-Softmax:
Latent-GRPO \citep{deng2026latentgrpo} uses top-$k$ vocabulary
mixtures with one-sided Gumbel margins, invalid-sample advantage
masking, and optimal-correct-path first-token selection;
SofT-GRPO \citep{zheng2026softgrpo} adds Gumbel noise to logits and
uses Gumbel reparameterisation to assign credit to the underlying
logits. Both methods rely on the latent being samplable \emph{by
construction}: their RL story requires a tractable density at every
latent position. Hidden-state-recurrence latents admit no such
density. Switch-GRPO (\S\ref{ssec:method-grpo}) extends GRPO to this
regime by defining the policy ratio only at text positions while
keeping rollouts that perform real hidden-state injection. This
complements the vocabulary-mixture line: we ask how far
hidden-state recurrence can be pushed, given that it is
weight-sharing with the LM and admits the simplest implementation.

\subsection{Interpretability of Internal Reasoning States}

A central concern with any latent-CoT method is whether the latent
states carry meaningful reasoning content. \textbf{Logit-lens}
analysis \citep{nostalgebraist2020logitlens,belrose2023eliciting}
reads intermediate hidden states through the LM head to obtain a
distribution over the vocabulary, offering a qualitative view of
what the model ``believes'' at each layer or step. \textbf{Linear
probing}
\citep{tenney2019bertrediscovers,belinkov2022probingsurvey} trains
a linear classifier on frozen activations to test whether a target
property is encoded in a particular layer. \textbf{Causal
activation interventions}
\citep{meng2022rome,heimersheim2024patching} perturb specific
activations and measure the effect on the output distribution,
turning correlational evidence into a causal claim. We use all
three in \S\ref{sec:mechanistic}; to our knowledge this is the
first study to apply this triad to a learned latent-CoT model at
$8$B scale.

\section{Extended Discussion}
\label{sec:appendix-discussion}

\paragraph{Hidden-state recurrence is RL-compatible.}
Recent latent-CoT work has argued that hidden-state-recurrence
latents cannot be optimised with on-policy RL and has switched to
samplable vocabulary mixtures
\citep{deng2026latentgrpo,zheng2026softgrpo,zhang2025softthinking,deng2026latentsft}.
Switch-GRPO is a constructive counterexample. The key observation
is that the GRPO policy ratio only needs a tractable density at
the \emph{decision points} of the policy, which are the
text-positions that emit \swi, \swiend and the visible answer.
Latent positions have deterministic embeddings given the
preceding text; they only need to be replayed identically at the
gradient pass. Combined with engineering for the multi-pass
forward (Appendix~\ref{sec:appendix-impl}), this lets us keep
Coconut's latent formulation while still getting the gradient
signal of GRPO.

\paragraph{What RL actually changes.}
The mechanistic analysis (\S\ref{sec:mechanistic}) shows that
Switch-GRPO does not erase the switch policy curriculum SFT
already learned. After SFT, $p(\swi)$ is $0.85$ at every
annotated boundary with very low entropy; after Switch-GRPO it
drops to $0.48$ with entropy $\sim\!0.5$, while the contrast to
the immediate neighbours stays at $\sim\!10^2$. The switch rate
halves and latent-conditional accuracy nearly doubles, so the
simplest reading, supported by the intervention result, is that
RL has shifted probability mass away from boundaries where
opening a latent block would not have helped and toward
boundaries where it does.

\paragraph{The latent step is not a ``black box''.}
A persistent worry about non-decoding ``thinking'' tokens
\citep{pfau2024letsthinkdotbydot,goyal2024pause} is that they
function as inert compute rather than as task-relevant reasoning
steps. \S\ref{ssec:mech-intervention} addresses this for
\method's latent blocks: on problems where the model uses latent
reasoning and answers correctly, zeroing the injected hidden
states costs $66.7$ accuracy points, while replacing them with a
random vector of the same norm costs only $9.5$ points. An
arbitrary non-zero perturbation does not reproduce the latent
computation; the specific hidden state Eq.~\ref{eq:hsi} produces
is what the answer depends on.

\end{document}

%% file: arxiv_lark/LARK_Lab_Arxiv_Template/math_commands.tex
\usepackage{amsmath,amsfonts,bm}

\def\eqref#1{equation~\ref{#1}}

\def\1{\bm{1}}

\DeclareMathAlphabet{\mathsfit}{\encodingdefault}{\sfdefault}{m}{sl}
\SetMathAlphabet{\mathsfit}{bold}{\encodingdefault}{\sfdefault}{bx}{n}

